\documentclass[conference]{IEEEtran}
\IEEEoverridecommandlockouts
\usepackage{cite}
\usepackage{amsmath,amssymb,amsfonts}
\usepackage{colortbl}
\usepackage{xcolor}
\usepackage{algorithmic}
\usepackage{graphicx}
\usepackage{textcomp}
\usepackage{xcolor}
\usepackage{caption}
\usepackage{subcaption}

\def\BibTeX{{\rm B\kern-.05em{\sc i\kern-.025em b}\kern-.08em
    T\kern-.1667em\lower.7ex\hbox{E}\kern-.125emX}}

\makeatletter
\def\ps@IEEEtitlepagestyle{%
\def\@oddfoot{\mycopyrightnotice}%
\def\@evenfoot{}%
}

\def\mycopyrightnotice{%
{\footnotesize 979-8-3315-4112-5/25/\$31.00~\copyright~2025 IEEE\hfill} 
\gdef\mycopyrightnotice{}
}

\begin{document}

\newcommand{\LOOPLC}{$\mathcal{LOOP-LC}$}

\title{ 
Impact of Data Poisoning Attacks on Feasibility and Optimality of Neural Power System Optimizers

\thanks{This material is based upon work supported by the National Science Foundation (NSF) Graduate Research Fellowship under Grant No. DGE 2137420. Any opinion, findings, and conclusions or recommendations expressed in this material are those of the authors(s) and do not necessarily reflect NSF views.}
}

\author{\IEEEauthorblockN{Nora Agah}
\IEEEauthorblockA{
\textit{The University of Texas at Austin}\\
Austin, Texas, USA \\
Norakagah@utexas.edu}
\and
\IEEEauthorblockN{Meiyi Li}
\IEEEauthorblockA{
\textit{The University of Texas at Austin}\\
Austin, Texas, USA \\
 meiyil@utexas.edu}
\and
\IEEEauthorblockN{Javad Mohammadi}
\IEEEauthorblockA{\textit{The University of Texas at Austin}\\
Austin, Texas, USA \\
javadm@utexas.edu}
}

\maketitle

\begin{abstract} 
The increased integration of clean yet stochastic energy resources and the growing number of extreme weather events are narrowing the decision-making window of power grid operators. This time constraint is fueling a plethora of research on Machine Learning (ML)-based optimization proxies. While finding a fast solution is appealing, the inherent vulnerabilities of the learning-based methods are hindering their adoption. One of these vulnerabilities is data poisoning attacks, which adds perturbations to ML training data, leading to incorrect decisions. The impact of poisoning attacks on learning-based power system optimizers have not been thoroughly studied, which creates a critical vulnerability. In this paper, we examine the impact of data poisoning attacks on ML-based optimization proxies that are used to solve the DC Optimal Power Flow problem. Specifically, we compare the resilience of three different methods—a penalty-based method, a post-repair approach, and a direct mapping approach—against the adverse effects of poisoning attacks. We will use the optimality and feasibility of these proxies as performance metrics. The insights of this work will establish a foundation for enhancing the resilience of neural power system optimizers.

\end{abstract}

\begin{IEEEkeywords}
Poisoning Attack, Machine Learning, Optimal Power Flow
\end{IEEEkeywords}

\vspace{-.2cm}
\section{Introduction}
\vspace{-.2cm}
\subsection{Motivation} 

Today's power grid is unprepared to face its operational challenges ranging from an ever--increasing integration of renewables to an unprecedented number of extreme weather events. These factors increase the power system's variability, necessitating rapid dispatch decision-making. Put differently, these complexities can derail the balance of supply and demand and make solving power dispatch problems, like Optimal Power Flow (OPF), even more challenging.

Power grid operators often need to solve the OPF problem every 15 minutes and they rely on traditional solvers, that utilize time--consuming iterative algorithms like \cite{matpow2011, wachter2006interior}, to accomplish this task. 
The traditional iterative processes, while reliable, struggle to keep up with the rapid changes and variability introduced by renewable integration and extreme weather events. Moreover, the OPF problem's significance in grid operation makes it a prime target for cyber-intrusions, which often aim to disrupt the delicate balance between supply and demand \cite{Duan2018DataInteg, Rahman2014OPFAtt, Yang2017DataInteg}. As the traditional methods become inadequate under these evolving challenges, the potential of Machine Learning (ML) becomes increasingly apparent. ML-based optimization proxies utilize historical data in an offline training phase to develop models (i.e., neural networks (NN)) that achieve rapid online computation. This approach offers a much faster alternative to traditional solvers, significantly speeding up the decision-making process in grid operations.


While ML-based optimization proxies present a promising tool for power system optimization, they introduce the risk of generating non-feasible solutions. Infeasibilities result in system failures and grid outages.  Traditional solvers, though slower, consistently provide optimal solutions that adhere to the physical constraints of the grid. To apply ML-based optimization proxies effectively, it is necessary to assess the risks (including poor performance) thoroughly. A significant concern is the vulnerability of optimization proxies to adversarial attacks \cite{szegedy2014intrig}, where minor perturbations to the input data can cause the neural network to generate decisions that deviate significantly from optimality and violate system constraints. For instance, optimization proxies often rely on data generated by traditional solvers (such as a MATPOWER solver\cite{matpow2011}) for training. Hence, any attack on the training data provided by these solvers would jeopardize the performance of the trained models. The result of the data corruption can be severe, including financial losses due to generators producing excess power, damage to transmission lines from unworkable network solutions, and even blackouts resulting from insufficient power production. Therefore, it is crucial to investigate the impacts of such attacks on the optimality and feasibility of decisions made by ML-based optimization proxies.

\subsection{Literature Review} 

Extensive research has been devoted to developing optimization proxies that expedite the resolution of the OPF problem while ensuring the feasibility of decisions. A commonly employed method is the incorporation of penalty terms aimed at minimizing constraint violations \cite{Liu2022Pen, pan2022deepopf, liu2024teaching}. Although this approach is applicable across a variety of power system optimization problems, it merely penalizes rather than eliminating violations. Alternatively, some methods introduce a repair module that utilizes iterative algorithms to adjust the solutions. These methods range from using commercial solvers to identify the closest feasible solution \cite{zhao2020deepopf}, to optimizing a function that minimizes violations \cite{donti2021dc3}.  Further advancements involve differentiable mapping functions that directly enforce constraint satisfaction \cite{ Li_2023, chen2023end}. For instance, in our previous work \cite{li2023learning}, we introduced an iteration-free optimization proxy that ensures the hard feasibility of solutions through a single feed-forward process. However, research into assessing these methods' performance, such as optimality and feasibility, particularly under adversarial conditions, remains sparse.

Adversarial attacks in the power system context typically involve introducing limited perturbations to input data, aiming to maximize disruption in the output of optimization proxies. These attacks are categorized into two primary types: evasion and poisoning attacks. Evasion attacks occur during the operational or test phase of the NN, where perturbations in the input data compromise the output by misleading the neural network. Such attacks have been extensively studied within power systems \cite{Wang2024Evasion}. In contrast, poisoning attacks happen during the training phase, where perturbations to the training data alter the NN's weights, potentially causing long-term disruptions. Poisoning attacks are less explored, partly due to the assumption that training data is more secure than operational data \cite{agah2024datapois}. However, poisoning attacks, unlike evasion attacks that affect only a single output session, can distort the structure of the neural network until retraining occurs, thus impacting all subsequent outputs significantly \cite{abbasi2023brainwash}.

The extensive impact of data poisoning attacks has also led to research on a variety of methods to strengthen the robustness of NNs against these attacks \cite{CompSurvey}. This research includes methods for detecting poisoned data \cite{DetectPois}, or training the network to be more resilient to poisoned data \cite{inbook}. Understanding data poisoning attacks in relation to the power system will allow these robustness strategies to be applied effectively, helping strengthen security of ML methods used in grid operation.

In this paper, we focus on today's common training practices and shed light on their data poisoning vulnerabilities.
Specifically given that many OPF optimization proxies utilize MATPOWER data transferred from MATLAB to Python for training \cite{zhao2020deepopf, donti2021dc3, Li_2023}, there exists a notable intrusion vulnerability. 
Thus, a thorough study of data poisoning is crucial for assessing the reliability of ML-based optimization proxies.

\subsection{Contribution} In this paper, we investigate the impact of data poisoning attacks on the optimality and feasibility of ML-based optimization proxies applied to solving the DC-OPF problem.
We conduct tests on three distinct types of ML-based optimization proxies: the penalty method \cite{Liu2022Pen}, the DC3 method \cite{Li_2023} (a post-repair approach), and the \LOOPLC~ method \cite{donti2021dc3} (a direct mapping approach). Our research aims to develop more robust ML methods that can withstand adversarial threats, thereby enhancing both the reliability and security of power system operations. In addition, our findings provide valuable insights for power system operators and can facilitate the adoption of neural-based optimizers. This paper is organized as follows: Section \ref{PF} describes the problem formulation for the DC-OPF problem, Section \ref{PA} details our implementation of the data poisoning attack, Section \ref{Res} discusses the results of the attack simulation, and finally Section \ref{conc} provides our conclusion and potential extensions of this work.

\section{Problem formulation}\label{PF}
\subsection{Optimal Power Flow}  

 The OPF is a fundamental optimization problem in power grid operations, aimed at finding the most cost-effective generation profiles while adhering to the physical constraints of the power grid. This work specifically focuses on the simplified version of the OPF problem, i.e., DC-OPF, where its formulation is presented below.

\small
\begin{subequations}
\begin{gather}
    \texttt{min} \sum_{i \in \mathcal{G}} C_i(P_{\texttt{G}}^{i}) \label{min} \\
    \texttt{subject to:} 
    \sum_{i \in \mathcal{G}_{k}} P_{\texttt{G}}^{i} - \sum_{j \in \mathcal{L}_{k}} P_{\texttt{D}}^{j} = \sum_{l \in \mathcal{B}_{k}} P_{l,k}, \quad \forall k \in \mathcal{N} \label{PB} \\
    P_{l,k} = \frac{\theta_{l} - \theta_{k}}{X_{l,k}}, \quad \forall (l, k) \in \mathcal{L} \label{LF} \\
    P_{\texttt{G}}^{\text{min},i} \leq P_{\texttt{G}}^{i} \leq P_{\texttt{G}}^{\text{max},i}, \quad \forall i \in \mathcal{G} \label{GL} \\
    \theta_{k}^{\text{min}} \leq \theta_{k} \leq \theta_{k}^{\text{max}}, \quad \forall k \in \mathcal{N} \label{PA} \\
    -P_{l,k}^{\text{max}} \leq P_{l,k} \leq P_{l,k}^{\text{max}}, \quad \forall (l, k) \in \mathcal{L} \label{LC}
\end{gather}
\end{subequations}
\normalsize

This problem aims to minimize the total generation cost, i.e., \( \sum_{i \in \mathcal{G}} C_i(P_{\texttt{G}}^{i}) \). In \eqref{min}, \( C_i(P_{\texttt{G}}^{i}) \) represents the generation cost of plant \( i \). The active power output of generator \( i \) is denoted by \( P_{\texttt{G}}^{i} \), and the set of all generators is given by \( \mathcal{G} \). The load at bus \( j \) is denoted by \( P_{\texttt{D}}^{j} \), and the set of loads is given by \( \mathcal{L} \).

The power balance constraint is captured by \eqref{PB} and ensures supply matches demand. Note, the power exchange of bus $i$ with neighboring buses is represented by \( \sum_{l \in \mathcal{B}_{k}} P_{l,k} \). The set of lines connected to bus \( k \) is shown by \( \mathcal{B}_{k} \).
%
The active power flow on line \( (l, k) \) is given by \( P_{l,k} \), and it is defined by \( P_{l,k} = \frac{\theta_{l} - \theta_{k}}{X_{l,k}} \), where \( \theta_{k} \) and \( \theta_{l} \) are the voltage phase angles at buses \( k \) and \( l \), respectively, and \( X_{l,k} \) is line's reactance.

Generators' output \( P_{\texttt{G}}^{i} \) is limited to lower and upper bounds, \( P_{\texttt{G}}^{\text{min},i} \) and \( P_{\texttt{G}}^{\text{max},i} \), respectively. Similarly, phase angle limits are enforced by (1e) for all buses. Finally, line capacity limits ensure that the power flow on line \( (l, k) \) stays within its limits.
%
The sets \( \mathcal{N} \) and \( \mathcal{L} \) represent the buses and loads in the network.



\subsection{ML-based Optimization Proxies}The OPF problem, as outlined in \eqref{min}-\eqref{LC}, is typically resolved every 15 minutes to accommodate updated load demands ($P_D$). Traditionally handled using iterative solvers, this repeated process can alternatively be managed using ML-based optimization proxies. These proxies leverage historical data to develop a mapping between demand and the optimal generation profile, enabling faster, real-time decision-making.

Most existing OPF optimization proxies, including those we examine in this paper, utilize MATPOWER data that is transferred from MATLAB to Python for training. This data transfer introduces a potential vulnerability, as it could be targeted by data corruption attacks. In the following section, we discuss how data poisoning attacks are implemented against three distinct types of optimization proxies.

\section{Poisoning Attack Implementation}\label{PA}

Poisoning attacks generally adopt one of two strategies. In a black box attack, the attacker has limited knowledge about NN and its data, relying mostly on estimations to inflict damage. Conversely, a white box attack assumes that an attacker has access to considerable details of the grid network and its data. In our study, we employ a white box attack approach, where the attacker is well-informed about the dataset and has a rough idea of the network's architecture. This enables us to demonstrate the significant impact on the optimal solutions produced by the method. Our approach modifies the strategy outlined by Chen et al. in their study of evasion attacks on demand prediction \cite{evasion2019demand}. We specifically engineer an attack to maximize the loss, misleading the neural network into predicting higher generation needs than required. This leads to an unnecessary increase in energy production, thus disrupting the supply-demand equilibrium.

In what follows, we will outline our methods of poisoning attacks.
Then, we will detail the implementation of poisoning attacks on three different OPF proxies: the penalty method \cite{Liu2022Pen}, the DC3 method \cite{Li_2023}, and the \LOOPLC~ method \cite{donti2021dc3}.

\subsection{Formulating the Poisoning Attack Strategy}Data poisoning attacks are meticulously crafted to manipulate the training data of ML-based optimization proxies. These proxies are designed to map load demands, the inputs, to their respective optimal generation profiles, the outputs. By subtly introducing calculated perturbations into the load demand data, attackers aim to skew the proxy's ability to predict the most efficient (optimal and feasible) power output setting.

The perturbations are guided by the gradient of a Mean Squared Error (MSE) loss function, which measures the deviation between the perturbed (attacked) generation targets and the proxy's predictions. The objective is to manipulate the model into recommending higher than necessary energy outputs, maximizing the loss \cite{bai2021recent}:

\small
\begin{equation}
\texttt{max}_\delta \; L(\theta, x + \delta, y)\label{pois}
\end{equation}
\normalsize

In \eqref{pois}, $L$ is the loss function, $\delta$ is the perturbation, $\theta$ is the model parameters, $x$ is the model input, and $y$ is the output.

A bounding mechanism is applied to increase the likelihood that these perturbations evade detection by protective measures. This mechanism restricts the degree of perturbation applied to each data point while showing {\color{black}pre-existing knowledge of
grid operation (shown by $x_{\text{orig}}$)}:


\vspace{-.4cm}

\small
\begin{equation}
x_{\text{new, bounded}} = \texttt{clip}\left(x_{\text{new}}, \, x_{\text{orig}} - \text{L} \cdot |x_{\text{orig}}|, \, x_{\text{orig}} + \text{L} \cdot |x_{\text{orig}}|\right) \label{clip}
\end{equation}
\normalsize

\vspace{.2cm}

The functionality of the clip function is summarized below:
\small
\begin{equation}
\texttt{clip}(x, a, b) = 
\begin{cases}
x = a, & \text{if } x < a \\
x = x, & \text{if } a \leq x \leq b \\
x = b, & \text{if } x > b
\end{cases}
\end{equation}
\normalsize

\noindent where $L$ is the bound that we select to avoid detection, $x_\text{new}$ is the new, perturbed input data that has been attacked. Also, $x_\text{orig}$ is the unattacked, original input data, and $x_\text{new, bounded}$ is the new, attacked data once the bound has been applied. This approach strategically enforces limits on the alterations to ensure the changes do not attract undue attention, while still achieving the intended disruptions to model training. This subtle yet strategic manipulation leads to optimality and feasibility gaps in the grid operation once the model is deployed, demonstrating the critical need for robust defenses against such adversarial threats in the training of ML-based optimization proxies. In the following proxy methods, we perturb the input data at the same location in the data pipeline, however, the way in which that perturbed data will affect the algorithm is different for each case.  \label{FormPoisAttack}

\subsection{Poisoning Optimization Proxies}In the following, we
discuss how data poisoning attacks are implemented against
three distinct types of optimization proxies:  the penalty method \cite{Liu2022Pen}, the DC3 method \cite{Li_2023} (a post--repair approach), and the \LOOPLC~ method \cite{donti2021dc3} (a direct mapping approach).

\subsubsection{Penalty Method}


The penalty method is a well-established approach used to manage feasibility violations in ML training. It incorporates constraints directly into the loss function, which is designed to increase as the NN outputs approach the boundaries of these constraints. This design incentivizes the network to produce solutions that remain within the feasible region.

\begin{figure}[h!]
\centerline{\includegraphics[scale=0.085]{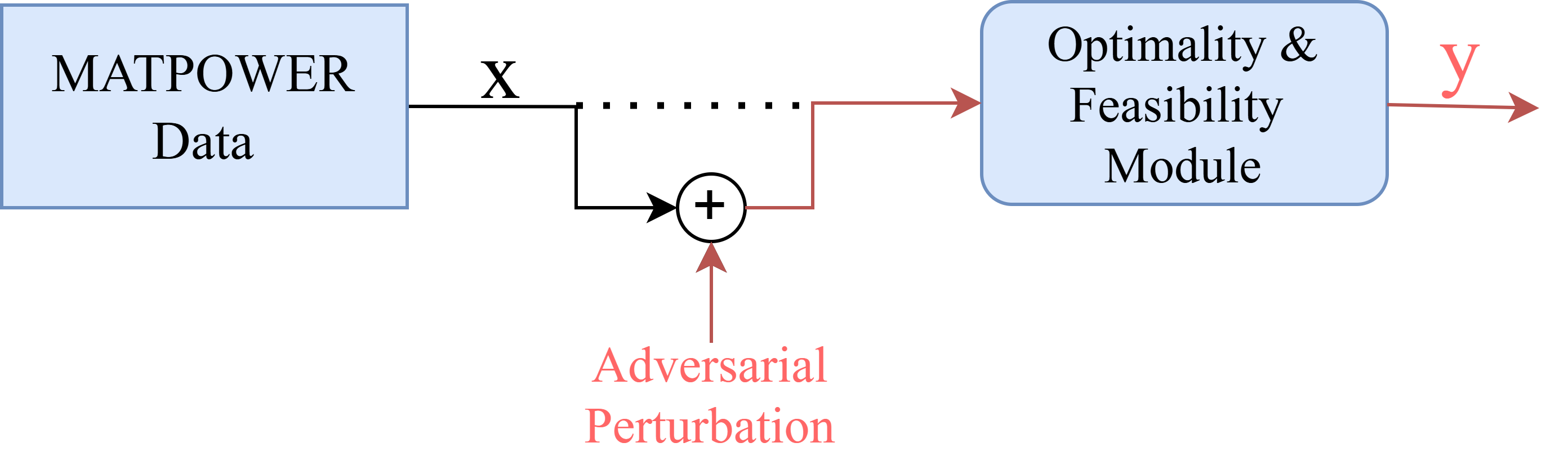}}
\vspace{-.2cm}
\caption{Poisoning the workflow of Penalty methods. 
}
\label{penfig}
\end{figure}

Our approach for introducing perturbations into the penalty method framework is shown in Fig. \ref{penfig}. By exploiting the inherent flexibility of penalty-based methods in enforcing feasibility constraints, adversarial perturbations can cause constraint violations and push solutions further from the optima.

\vspace{-.2cm}

\subsubsection{Deep Constraint Completion and Correction (DC3) Method} We also focused on poisoning the DC3 method from Donti et al in \cite{donti2021dc3}. This paper proposes a method to incorporate
feasibility constraints in deep learning in order to make
the process more usable for problems with real physical constraints like OPF, called Deep Constraint Completion and Correction (DC3). In \cite{donti2021dc3}, authors first solve the necessary equality constraints, then infer the remaining ones. To this end, they will
use deep NN to find a partial solution and later infer the remaining solutions using equality constraints. The resulting solution is then moved to the region that also
satisfies the inequality constraints (the feasible region) by taking gradient descent steps along the equality-satisfying region,
and the loss is back-propagated and training continues. As this
paper mentions, convergence to the optima is not guaranteed
in the gradient descent portion, but during test, the proposed
solution should be close to the actual solution. 

\begin{figure}[ht]
\centerline{\includegraphics[scale=0.059]{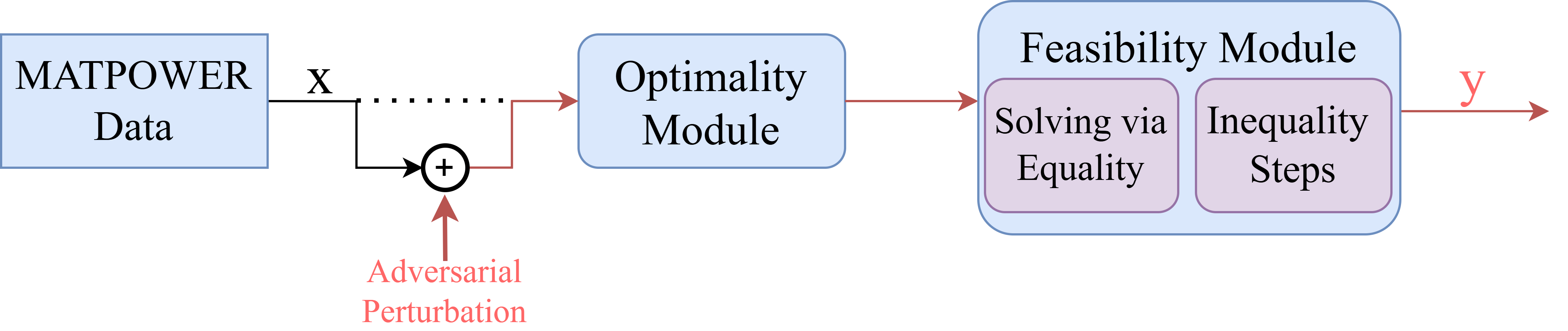}}
\vspace{-.2cm}
\caption{Poisoning the workflow of the DC3 method.}
\label{dc3fig}
\end{figure}

We expect a trade off between being close to the optimal value and being within the feasible region due to the walk into the feasible region potentially taking the solution further away from the optimal. The poisoning attack would exploit this inherent weakness by either causing a feasibility violation or keeping the solution far away from the true optima (see Fig. \ref{dc3fig}). In \cite{donti2021dc3}, the number of correction steps for convex problems is selected as ten to balance time, optimality, and feasibility. 

\subsubsection{Learning to Optimize the Optimization Process with Linear Constraints (LOOP--LC) Method} 


The \LOOPLC~method employs a neural approximator to transform input data directly into a near-optimal feasible solution. This approach ensures hard feasibility without relying on penalty terms and operates without iterations, utilizing a straightforward feed-forward process. The technique begins by training a neural network to map inputs to a virtual optimal point within the $\ell_\infty$-norm unit ball. Following this, a feasibility module is used to project this virtual prediction onto actual feasible solutions.

The method implements inequalities by using a gauge mapping technique to project the virtual optimal point from the \(\ell_\infty\)-norm unit ball onto the feasible space. For equality constraints, it determines the values of dependent variables based on the independent variables, effectively reconstructing the solution to the original problem. This integrated approach allows the \LOOPLC~method to achieve reliable and efficient optimization outcomes, free from the iterative adjustments typical of other methods.

\begin{figure}[h!]
\centerline{\includegraphics[scale=0.067]{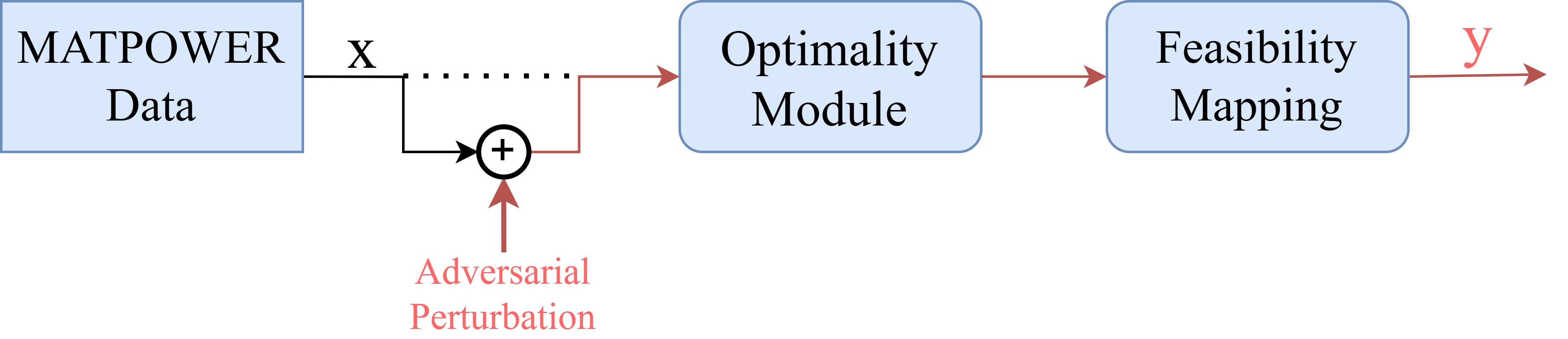}}
\vspace{-.2cm}
\caption{Poisoning the workflow of the \LOOPLC~method.}
\label{loopfig}
\end{figure}

 In the \LOOPLC~method, solution feasibility is guaranteed. Consequently, our poisoning attack strategy focuses on undermining the optimality of the solutions. We aim to shift the neural network's outputted solution further from the true optimal. This targeted manipulation occurs specifically within the ``Optimization module," as illustrated in Fig. \ref{loopfig}.

\captionsetup{skip=5pt} 

\begin{figure*}[h!]
\centering
\begin{subfigure}{0.45\linewidth}
    \includegraphics[width=\linewidth]{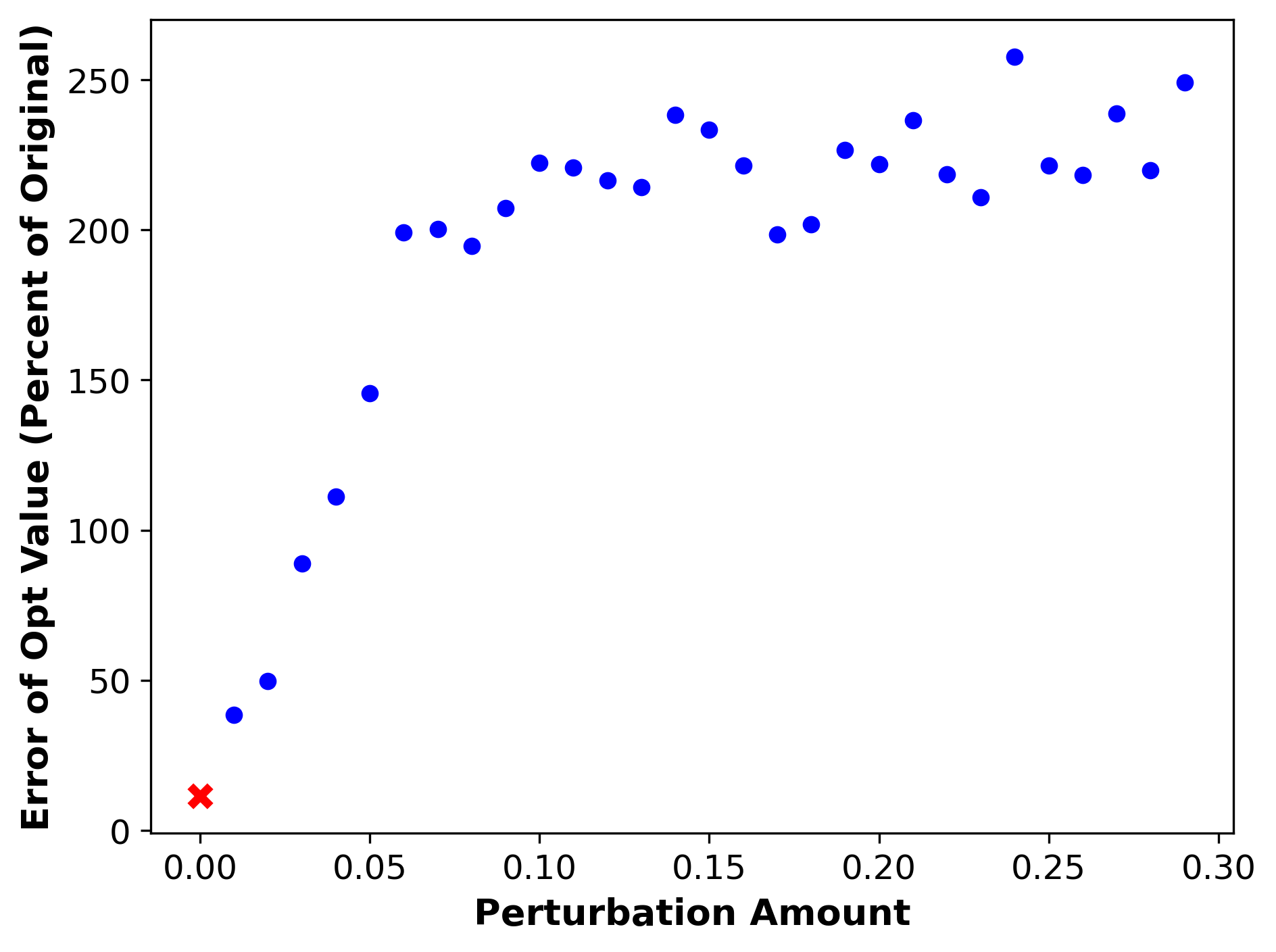}
    \caption{Penalty (Optimality) Error}
\end{subfigure}\hfill
\begin{subfigure}{0.45\linewidth}
    \includegraphics[width=\linewidth]{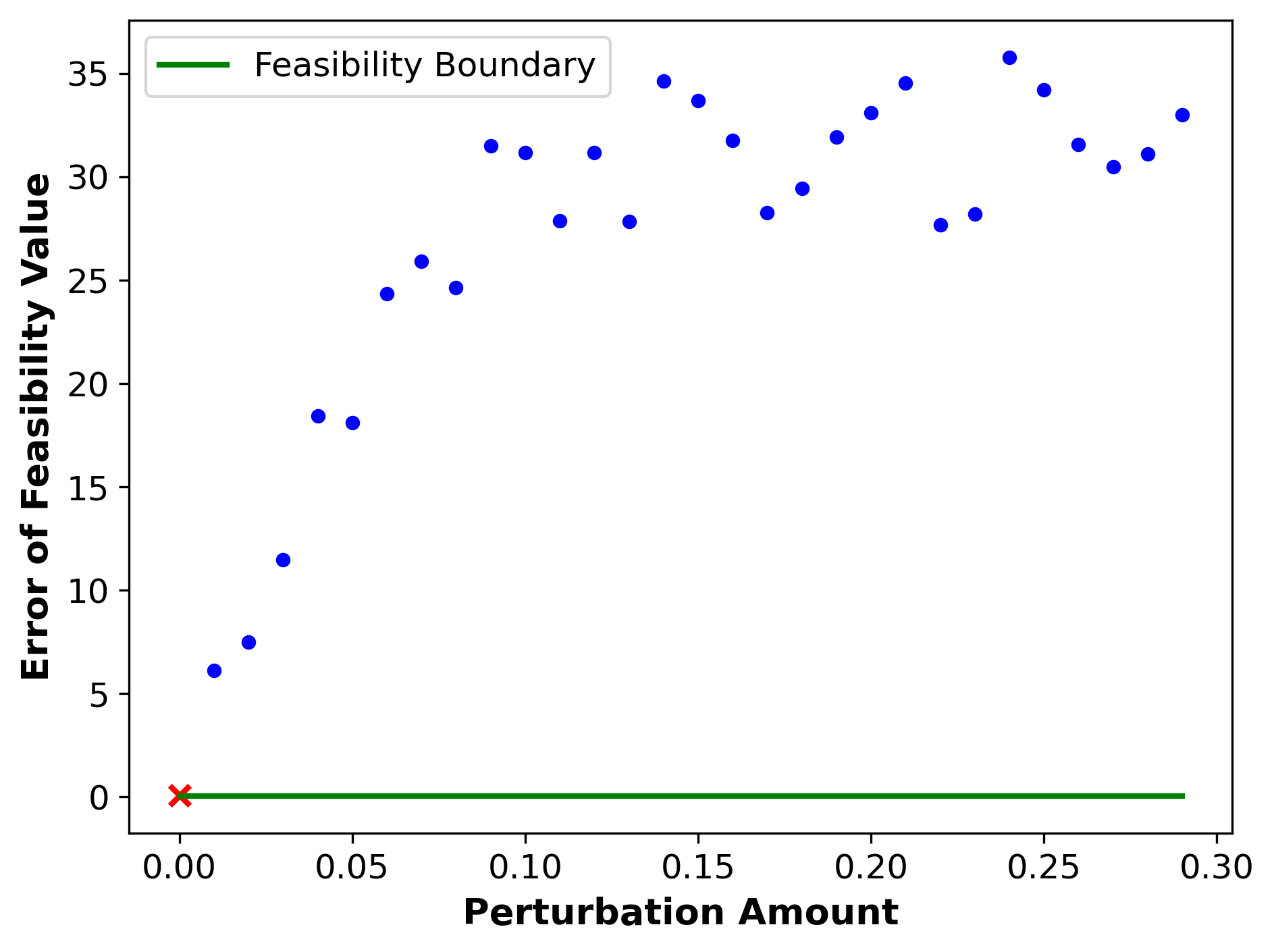}
    \caption{Penalty (Feasibility) Error}
\end{subfigure}

\vspace{10pt} 

\begin{subfigure}{0.45\linewidth}
    \includegraphics[width=\linewidth]{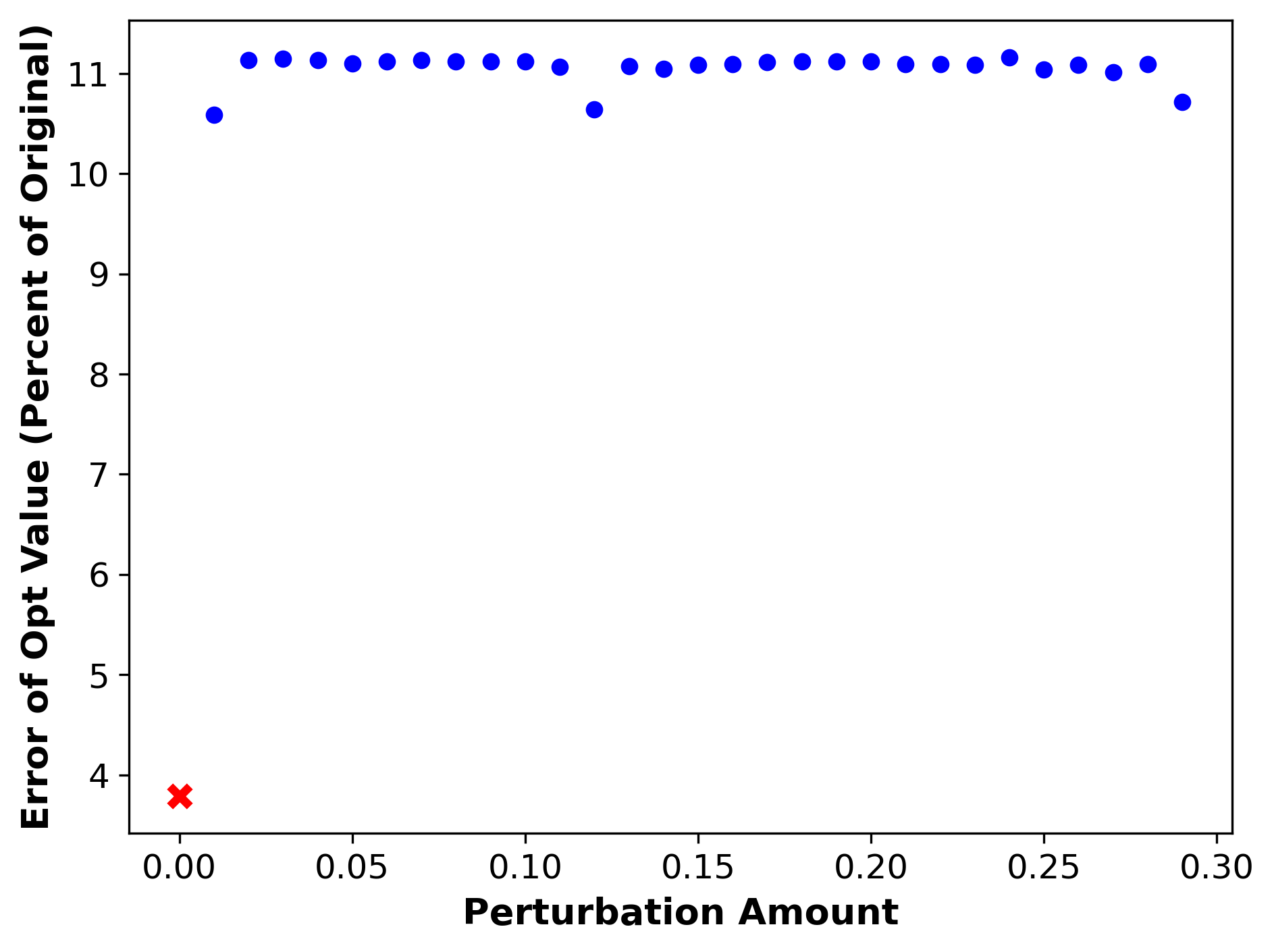}
    \caption{\LOOPLC~(Optimality) Error}
\end{subfigure}\hfill
\begin{subfigure}{0.45\linewidth}
    \includegraphics[width=\linewidth]{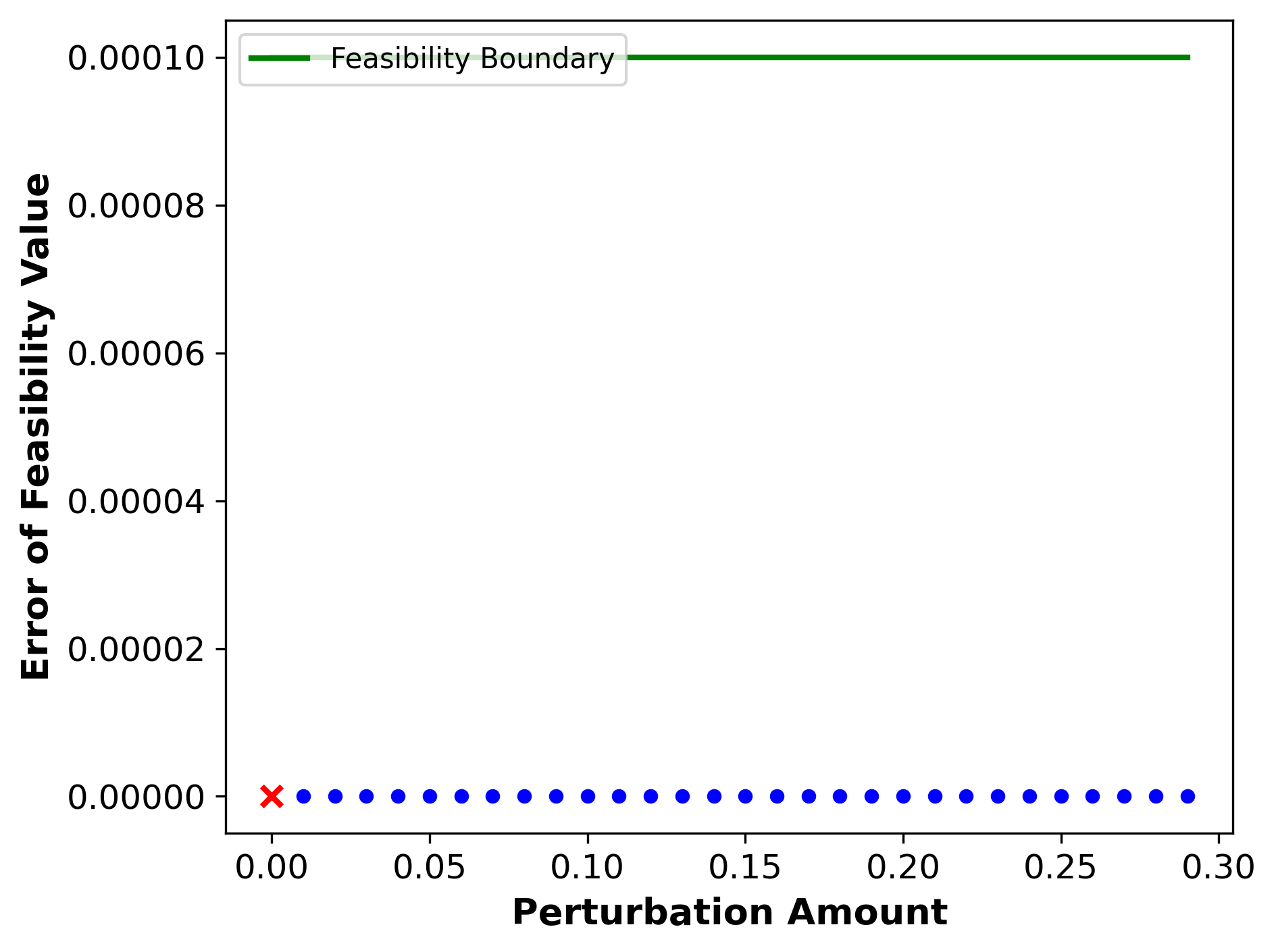}
    \caption{\LOOPLC~(Feasibility) Error}
\end{subfigure}

\vspace{10pt} 

\begin{subfigure}{0.45\linewidth}
    \includegraphics[width=\linewidth]{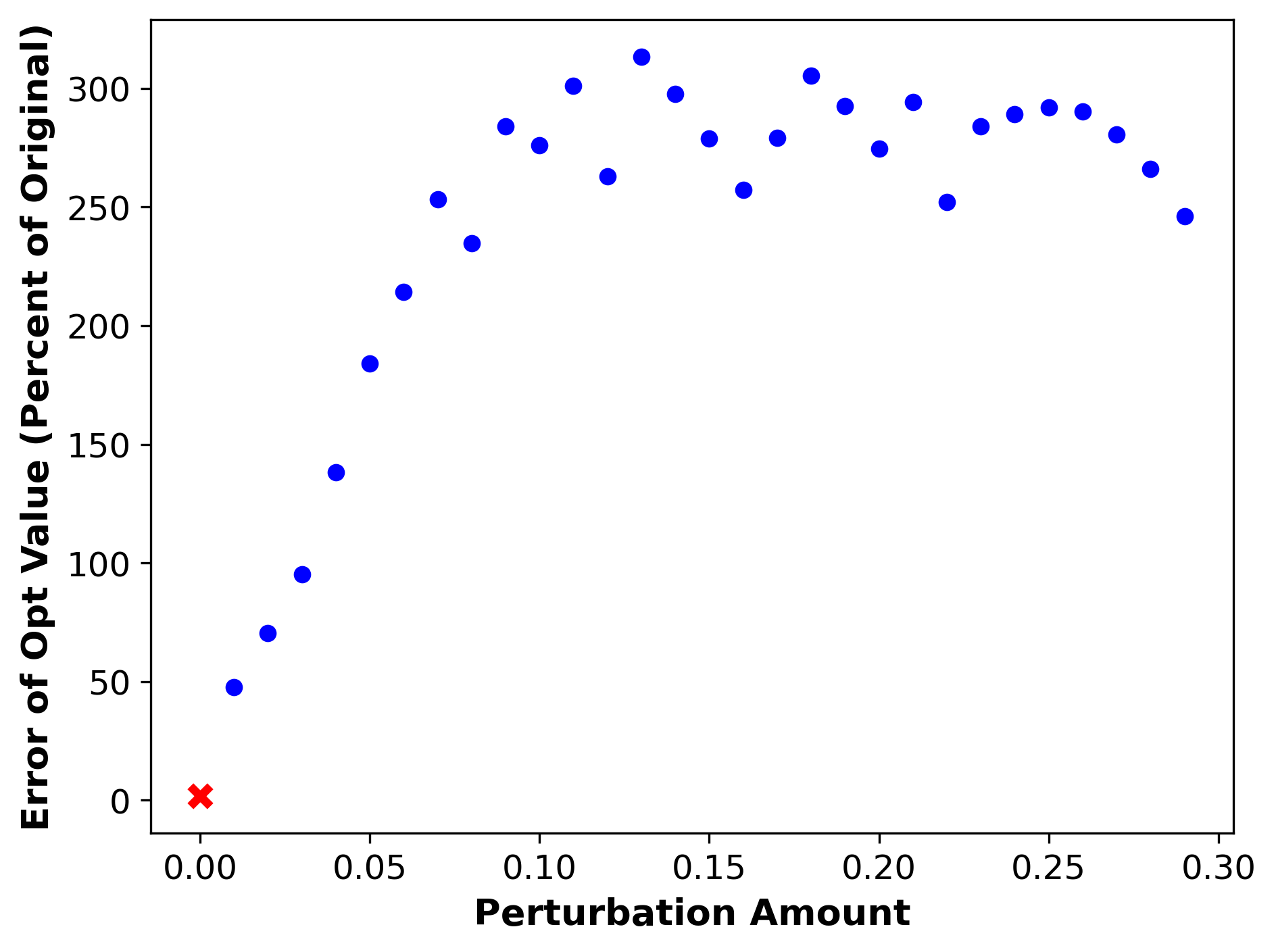}
    \caption{DC3 (Optimality) Error}
\end{subfigure}\hfill
\begin{subfigure}{0.45\linewidth}
    \includegraphics[width=\linewidth]{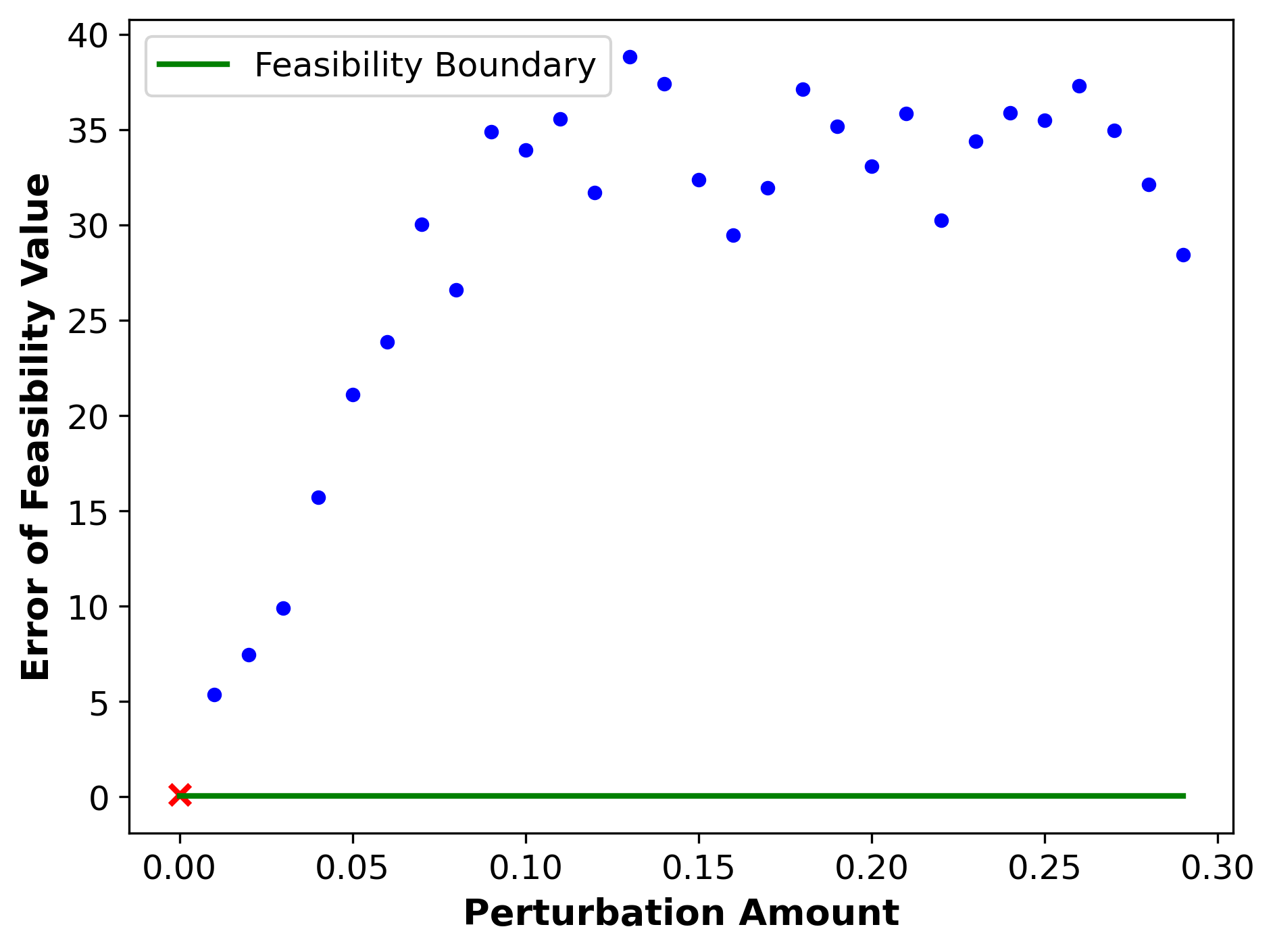}
    \caption{DC3 (Feasibility) Error}
\end{subfigure}

\caption{Plots of the error for each method (Penalty, \LOOPLC, and DC3) for both (left) optimality and (right) feasibility, versus perturbation amount, with a bound of 0.75. The values above the green feasibility line are infeasible. The red ``x" represents the unperturbed case. In the Penalty and DC3 methods, all results other than the unperturbed case are infeasible, while \LOOPLC~ maintains feasibility across all cases.}
\label{fig:all_methods}
\vspace{-0.3cm}
\end{figure*}

\begin{table*}[h!]
\caption{Summary of Adversarial Attack Results (Used on a perturbation bound of 0.75 $\times$ original value.)}
\vspace{-.2cm}
\begin{center}
\begin{tabular}{|c|>{\columncolor{green!20}}c|>{\columncolor{red!20}}c|>{\columncolor{green!20}}c|>{\columncolor{red!20}}c|>{\columncolor{green!20}}c|>{\columncolor{red!20}}c|}
\hline
 & \multicolumn{2}{c|}{\textbf{Optimality Gap}} & \multicolumn{2}{c|}{\textbf{Feasibility Gap}} & \multicolumn{2}{c|}{\textbf{Time on CPU (sec)}} \\

\hline 
\textbf{Optimizer} & \textbf{Unattacked} & \textbf{Attacked} & \textbf{Unattacked} & \textbf{Attacked} & \textbf{Unattacked} & \textbf{Attacked} \\
\hline 
\textbf{Penalty \cite{Liu2022Pen}} & 0.11412 & 2.20422 & 0.06166 & 29.95326 & 0.00017 & 0.00027\\
\hline
\textbf{LOOP-LC \cite{Li_2023} } & 0.03786 & 0.10964 & 0 & 0 & 0.06986 & 0.06801\\
\hline 
\textbf{DC3 \cite{donti2021dc3}} & 0.01654 & 2.77317 & 0.10314 & 33.25656 & 0.08053 & 0.10336\\
\hline 
\end{tabular}
\label{tab1}
\end{center}
\end{table*}

\vspace{-.2cm}
\section{Results} \label{Res}
\vspace{-.2cm}
\subsection{Simulation Setup}
\subsubsection{Dataset}
In order to compare the impact of the poisoning attack across methods, we used the same number of data points for all three methods, which was 200 samples of the IEEE 200 bus case generated from MATPOWER \cite{matpow2011}, as it is the way most of these methods generated their data. The input data consists of the real demand power, as it is for a DC problem.

\subsubsection{Metrics}
The optimality error (gap) is given by the difference between the true optimal value and the output of the attacked NN, divided by the true optimal value. We measure this gap for every output point and compute the average, and this is what we label as "Error of Opt Value" in Fig. \ref{fig:all_methods} and as the ``Optimality Gap" in Table \ref{tab1}. 
The feasibility error is given by the smallest distance between the output points and the feasible region. We find this for every output point and compute the average, and this is what we label as ``Feasibility of Opt Value" in Fig. \ref{fig:all_methods} and the ``Feasibility Gap" in Table \ref{tab1}.
\subsubsection{Motivation for The Choice of Bound}As discussed in Section \ref{FormPoisAttack}, to perturb the values, we add a perturbation amount, $\delta$, in the direction that maximizes loss. Once all the input points reach their bounded values, the error stays the same, regardless of the change in perturbation amount. This is seen in Fig. \ref{fig:all_methods},  where the error for both optimality and feasibility stops increasing once the perturbation added reaches the bound for the input points. Due to this, we focused on our choice of bound in our implementation and selected $\delta$ to reach the bound, as we wanted to demonstrate the extent of the damage that could be done by the attack. 

To select the bound, we considered the varying knowledge of multiple grid operators looking at the data. We assumed that the operators had a good intuition of the range of values to expect from the demand and generation. If values were set to zero or doubled, which would happen with a bound of 1, that would be apparent to operators. However, we wanted to show the extent of the attack's impact by suggesting a bound that was close enough to 1. The bound value of 0.75 was deemed reasonable, as the input values were formatted such that most were very small (much less than 1), and the bound of 0.75 was difficult to detect given our data set.

\subsection{Simulation Results and Discussions}Table \ref{tab1} summarizes the results of the poisoning attack on the different methods. As presented in the table, the poisoning attack has a significant impact on the optimality value for all three methods. For the two methods in which feasibility is not guaranteed (i.e., the penalty and DC3 methods) there is also a considerable impact on the solution feasibility.

It is insightful to compare the architectural differences and how they affect the results. For instance, DC3 shows considerable feasibility violations, with an average violation of about 33 per output point. With more steps walking to the feasible region, this may improve, but the time would increase.  Attacking the penalty-based method has a slightly smaller impact on its feasibility, an average violation of about 30 per output point, which is to be expected due to the penalty term not being strictly enforced. These feasibility violations are problematic since if these methods were implemented in the grid operation, these violations could have potentially adverse consequences, such as causing an imbalance in supply and demand. The \LOOPLC~shows no feasibility violations, as it prioritizes feasibility over optimality. 

Interestingly, \LOOPLC~shows a strong performance in the optimality context. This is most likely due to the back propagation of the error in order to correct the optimality that takes place after applying the feasibility module. However, even the optimality gap of \LOOPLC~was three times worse when the attack was implemented. This directly affects what the cost of electricity generation would be if an attack were to happen to grid data.

Regardless of the structure, data poisoning attacks had a negative effect on the outputs of the optimization proxies. It would appear that the most resilient optimization proxies need hard enforcement of the feasibility, in addition to taking errors into consideration during the training process. It is likely that factoring in other methods of adversarial robustness in addition to this type of structure would help ensure resilience in the face of malicious actors who intend to harm grid operations.

For example, the penalty has a smaller impact on its feasibility than the DC3 method, but with more steps DC3, could have a feasible result. This is a trade-off that increases the time and has a risk of moving the output further away from the optimal, as is seen by the greater time taken and greater optimality error in DC3 than the penalty method.




\section{Conclusion and Future Works}\label{conc} 
The impact of poisoning attacks on neural power system optimizers has been understudied compared to that of evasion attacks. Through this work, we have focused on this gap by implementing poisoning attacks on the latest ML--based optimization proxies focused on the Optimal Power Flow problem. We measured the adverse impact of data poisoning attacks on the optimality and feasibility of the solutions output by ML-based OPF solvers. We have also identified the impacts of different ML-based solver structures and their resilience to poisoning attacks.

We aim to expand this work on multiple fronts. First, we plan to implement poisoning attacks for AC versions of the different ML--based methods. It would also be beneficial to determine the impact of the Poisoning Attack on other methods with different structures or even with some adversarial samples used in training. Additionally, we will implement classical methods to improve the robustness of NN against adversarial attacks, such as adversarial training. 


\bibliographystyle{IEEEtran}
\bibliography{TPEC22}

\end{document}